\documentclass[
]{ceurart}

\usepackage{listings}
\usepackage[ruled,vlined,linesnumbered]{algorithm2e}
\usepackage{pifont}
\SetKwInput{KwInput}{Input}
\SetKwInput{KwOutput}{Output}
\definecolor{Skyblue}{RGB}{60,150,210}
\begin{document}

\copyrightyear{2025}
\copyrightclause{Copyright for this paper by its authors.
  Use permitted under Creative Commons License Attribution 4.0
  International (CC BY 4.0).}

\conference{GENNEXT@SIGIR'25: The 1st Workshop on Next Generation of IR and Recommender Systems with Language Agents, Generative Models, and Conversational AI, Jul 17, 2025, Padova, Italy}

\title{LLM-based User Profile Management for Recommender System}

\tnotemark[1]
% \tnotetext[1]{You can use this document as the template for preparing your
%   publication. We recommend using the latest version of the ceurart style.}

\author[1]{Seunghwan Bang}[%
orcid=0009-0005-3346-149X,
email=shbang1422@unist.ac.kr,
url=https://shwanbang.github.io/,
]
\address[1]{Ulsan National Institute of Science and Technology (UNIST), 50 UNIST-gil, Eonyang-eup, Ulju-gun, Ulsan, 44919, Republic of Korea}

\author[2]{Hwanjun Song}[%
orcid=0000-0002-1105-0818,
email=songhwanjun@kaist.ac.kr,
url=https://songhwanjun.github.io/,
]
\cormark[1]
\address[2]{Korea Advanced Institute of Science and Technology (KAIST), 291 Daehak-ro, Yuseong-gu, Daejeon, 34141, Republic of Korea}

\cortext[1]{Corresponding author.}

\begin{abstract}
The rapid advancement of Large Language Models (LLMs) has opened new opportunities in recommender systems by enabling recommendations without conventional training. Despite their potential, many existing works rely solely on users' purchase histories, leaving significant room for improvement by incorporating user-generated textual data, such as reviews and product descriptions. Addressing this gap, we propose PURE, a novel LLM-based recommendation framework that builds and maintains evolving user profiles by systematically extracting and summarizing key information from user reviews. PURE consists of three core components: a Review Extractor for identifying user preferences and key product features, a Profile Updater for refining and updating user profiles, and a Recommender for generating personalized recommendations using the most current profile. To evaluate PURE, we introduce a continuous sequential recommendation task that reflects real-world scenarios by adding reviews over time and updating predictions incrementally. Our experimental results on Amazon datasets demonstrate that PURE outperforms existing LLM-based methods, effectively leveraging long-term user information while managing token limitations.
\end{abstract}

\begin{keywords}
  Large Language Models (LLMs),
  Recommender Systems (RS),
  Personalization
\end{keywords}

\maketitle

\section{Introduction}

The rapid advancement of Large Language Models (LLMs) \cite{touvron2023llama, dubey2024llama, achiam2023gpt, team2024gemma} has significantly impacted various domains, such as text summarization \cite{lewis2019bart} and search~\cite{karpukhin2020dense}. Recent studies leverage LLMs in recommender systems for their human-like reasoning and external knowledge integration through in-context learning \cite{brown2020language} and retrieval-augmented generation \cite{lewis2020retrieval}. As such, LLMs exhibit the potential to be used as \emph{train-free} recommendation models without conventional training, which traditionally relies on explicit user-item interactions and training data \cite{he2017neural, kang2018self, he2020lightgcn}. 

Despite the advanced capability of LLMs, most recent works \cite{hou2024large, wei2024llmrec, ren2024representation, he2023large, zhai2023knowledge} rely solely on users' past purchase history (ie list of purchased items). This leaves significant room for further improvement by incorporating additional user-generated textual information, such as user reviews and product descriptions, which have yet to be fully leveraged.
In other words, they still fail to fully leverage various text data due to their inability to retain and process the increasing contextual information as users continue to make purchases, leading to longer recommendation sessions. This issue is primarily attributed to the \emph{omission} of the context, either due to the information loss within the LLM's memory \cite{liu2024lost} or the memory capacity by the token limit \cite{li2024survey, ding2024longrope}. 
Thus, extracting key features from a users' textual sources is essential, as demonstrated in MemoryBank~\cite{zhong2024memorybank}, a framework that enhances LLMs with \emph{long-term} memory by summarizing key information from conversations and updating user profiles. 

\begin{figure}
    \centering
    \includegraphics[width=\linewidth]{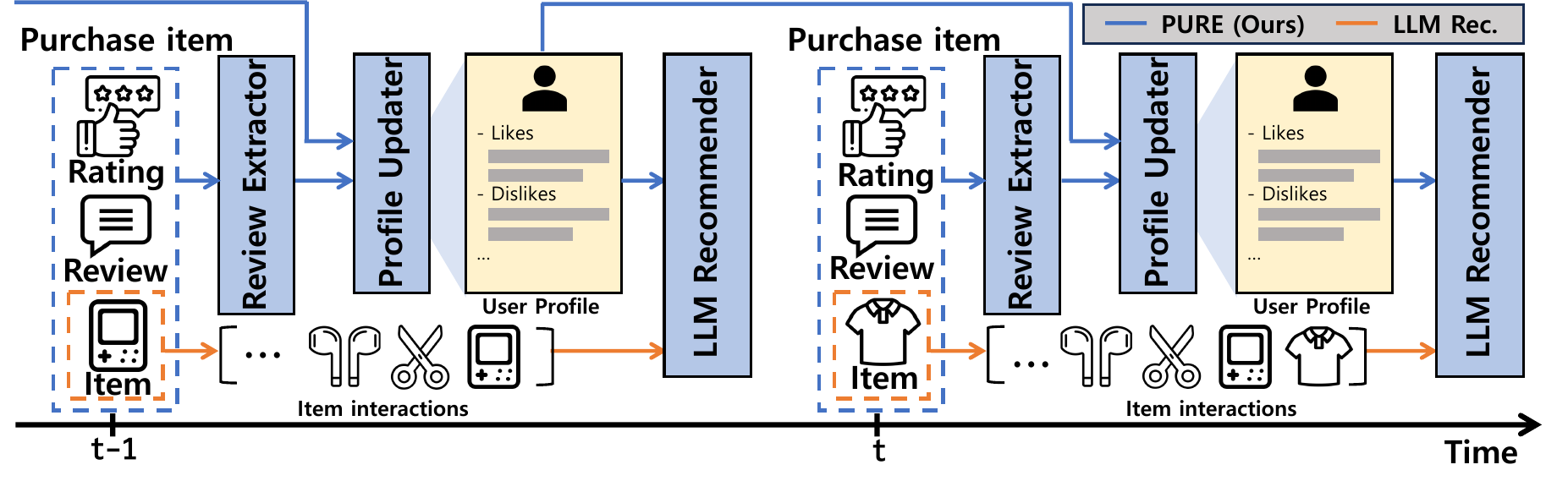}
    \caption{\textbf{Overall system of PURE.} \textcolor{Blue}{PURE} incorporates reviews, ratings, and item interactions, whereas \textcolor{Orange}{LLM Recommender} handles only item interactions. By using the "\textit{Review Extractor}" to identify key information and the "\textit{Profile Updater}" to refine the user profile, PURE addresses scalability issue (ie growth of input token size).}
    \label{fig:intro}
\end{figure}

Building on this foundation, we take the first step in extending LLMs' long-term memory beyond conversations in MemoryBank, adapting it to the evolving dynamics of recommendation systems.
We propose PURE, a novel LLM-based \underline{\textbf{P}}rofile \underline{\textbf{U}}pdate for \underline{\textbf{RE}}commender that constructs a user profile by integrating users' purchase history and user-generated reviews, which naturally expand as the recommendation sessions progress. As illustrated in ~\autoref{fig:intro}, PURE systematically extracts user likes, dislikes, and key features from reviews and integrates them into structured, dynamic user profiles.
Specifically, PURE consists of three main components: \underline{\emph{"Review Extractor"}}, which analyzes user reviews to identify and extract user likes, dislikes, and preferred product features, referred to as "key features", offering a comprehensive view of user interests and purchase-driving attributes; \underline{\emph{"Profile Updater"}}, which refines newly extracted representations by eliminating redundancies and resolves conflicts with the existing user profile, ensuring a compact and coherent user profile; and \underline{\emph{"Recommender"}}, which utilizes the most up-to-date user profile for recommendation task.

Our main contributions are as follows: 
(1) We propose PURE, a novel framework that systematically extracts and stores key information from user reviews, optimizing LLM memory management for the recommendation.
(2) We validate the effectiveness of PURE by introducing a more realistic sequential recommendation setting, where reviews are incrementally added over time, allowing the model to update user profiles and predict the next purchase continuously. This setup more accurately reflects real-world recommendation scenarios compared to prior works, which assume all past purchases are provided at once, ignoring the evolving nature of user preferences. 
(3) We empirically show that PURE surpasses existing LLM-based recommendation methods on Amazon data, demonstrating its effectiveness in leveraging lengthy purchase history and user reviews.

\section{Method}
\subsection{Problem Formulation}

In our recommender system, we consider the user $u$ dataset as follow: $\mathfrak{D}_u = \{\mathbf{R}_u, \mathbf{I}_u \}$, where $\mathbf{R}_u = \{r_u^1, \cdot\cdot\cdot, r_u^{k_u}\}$ represents the historical reviews, $\mathbf{I}_u = \{i_u^1, \cdot\cdot\cdot, i_u^{k_u} \}$ denotes the corresponding purchased items, and $k_u$ is the total number of purchased items from user $u$. 
Leveraging the user's dataset $\mathfrak{D}_u$, we aim to predict the next purchased item $i_u^{k_u+1}$ from a candidate set $\mathcal{C}_u^{k_u+1}$, which contains the ground-truth item. 

\smallskip
\noindent \textbf{One-shot Sequential Recommendation.} It predicts a single next item based on a static history of user interactions up to timestep $k_u-1$. Given the dataset $\mathfrak{D}_u$, the model observes $\mathfrak{D}_u^{k_u-1} = \{\mathbf{R}_u^{k_u-1}, \mathbf{I}_u^{k_u-1}\}$ and predicts the last item $i_u^{k_u}$ from the candidate set $\mathcal{C}_u^{k_u}$. This focuses on a one-time prediction without considering future timesteps.

\smallskip
\noindent \textbf{Continuous Sequential Recommendation.} This setup predicts the next item at every timestep ($4 \le t \le k_u-1$), making it a multi-step prediction task. At each timestep $t$, the model observes the updated interaction history $\mathfrak{D}_u^t = \{\mathbf{R}_u^t, \mathbf{I}_u^t\}$ and predicts the next item $i_u^{t+1}$ from the candidate set $\mathcal{C}_u^{t+1}$. This multi-step prediction process effectively captures temporal dependencies and allows continuous updates of user preferences, making it more aligned with real-world scenarios.

\subsection{PURE: \underline{P}rofile \underline{U}pdate for \underline{RE}commender}
In this section, we introduce PURE, novel framework that manages the user profile $\mathbf{P}_u$ from user reviews $\mathbf{R_u}$ and predict the next item with user profile. Algorithm~\ref{alg:main} can be divided into three steps.

\begingroup
\setlength{\textfloatsep}{8pt}

\begin{algorithm}
\DontPrintSemicolon
\SetAlgoLined
\SetNoFillComment
\LinesNotNumbered
\caption{PURE}
\label{alg:main}

\KwInput{
\small
Review extractor $\mathcal{E}(\cdot)$,
User profile updater $\mathcal{U}(\cdot)$,\\
\hspace*{3.5em} Recommender $\mathcal{R}(\cdot)$,
Dataset $\mathfrak{D}_u = \{\mathbf{R}_u, \mathbf{I}_u\}$ for user $u$,\\
\hspace*{3.5em} User profile $\mathbf{P}_u^t$, next purchase candidates $\mathcal{C}_u^{t+1}$, timestep $t$
}

{\small{\textcolor{Skyblue}{\# Extract representations from reviews}}}\\
$\tilde{l}_u^t, \tilde{d}_u^t, \tilde{f}_u^t = \mathcal{E}(r_u^t)$\\
$\hat{l}_u^t = l_u^{t-1} \cup \tilde{l}_u^t$ \quad{\small{\textcolor{Skyblue}{$\triangleright$ List of items user likes}}}\\
$\hat{d}_u^t = d_u^{t-1} \cup \tilde{d}_u^t$ \quad{\small{\textcolor{Skyblue}{$\triangleright$ List of items user dislikes}}}\\
$\hat{f}_u^t = f_u^{t-1} \cup \tilde{f}_u^t$ \quad{\small{\textcolor{Skyblue}{$\triangleright$ List of user's key features}}}\\

{\small{\textcolor{Skyblue}{\# Update user profile after redundancy removal}}}\\
$l_u^t, d_u^t, f_u^t = \mathcal{U}(\hat{l}_u^t, \hat{d}_u^t, \hat{f}_u^t)$\\
$\mathbf{P}_u^t = \{l_u^t, d_u^t, f_u^t\}$\\

{\small{\textcolor{Skyblue}{\# Recommend next purchase item}}}\\
$\mathrm{pred} = \mathcal{R}(\mathbf{P}_u^t, \mathbf{I}_u^t, \mathcal{C}_u^{t+1})$\\

\KwOutput{$\mathrm{pred}$}

\end{algorithm}
\endgroup

\smallskip\smallskip
\noindent \textbf{STEP 1: Extract User Representation. }
\noindent We begin by providing the LLM with raw inputs, including user reviews $\mathbf{R_u}$ and product names $\mathbf{I_u}$. The LLM extracts $\tilde{l_u^t}$(items the user likes), $\tilde{d_u^t}$(items the user dislikes), and $\tilde{f_u^t}$(key user features) from the incoming review as user representation. To do so, we utilized the following prompt template:

\smallskip\smallskip
\noindent\texttt{I purchased the following products and left reviews in chronological order: \{Asins, product names, input reviews\}. Analyze user's likes/dislikes/key features by referring to their reviews.}

\smallskip\smallskip
\noindent \textbf{STEP 2: Update User Profile. }
\noindent After the extraction in STEP 1, the extracted representation <$\tilde{l_u^t}$, $\tilde{d_u^t}$, $\tilde{f_u^t}$> concatenates with previous user profile $\mathbf{P}_u^{t-1} = \{l_u^{t-1}, d_u^{t-1}, f_u^{t-1} \}$. However, this faces a scalability issue as the number of reviews increases. Thus, leveraging the previous profile, we use an LLM to remove redundant and conflicting content from the extracted representation, yielding a more compact and up-to-date user profile $\mathbf{P}_u^t$ after concatenation. To achieve this, we utilized the following prompt template: 

\smallskip\smallskip
\noindent\texttt{You are given a list: \{list\}. Update this list by removing redundant or overlapping information. Note that crucial information should be preserved.}

\smallskip\smallskip
\noindent \textbf{STEP 3: Recommend Next Purchase Item. }
\noindent Recommender $\mathcal{R}$ reranks the given candidate item list to predict the user's next purchase by leveraging the updated profile $\mathbf{P}_u^t$ and purchased items $\mathbf{I}_u$. As such, here is the prompt template that we utilized: 

\smallskip
\noindent\texttt{Positive aspects: \{likes\}}
\noindent\texttt{Negative aspects: \{dislikes\}}
\noindent\texttt{Key Features: \{key features\}}
\noindent\texttt{Based on these inputs, rank the \{candidate list\} from 1 to 20 by evaluating their likelihood of being purchased.}

\section{Experiment}
\textbf{Datasets.} For a thorough evaluation, we utilize two datasets from the Amazon collection~\cite{ni2019justifying}: Video Games and Movies \&\ TV. To ensure a comprehensive analysis, we select datasets with diverse statistical properties, particularly in terms of the number of items. Each dataset includes ASINs, product names, and user reviews, which are chronologically sorted per user to reflect real-world behavior.

\smallskip\smallskip
\noindent \textbf{Baselines.} 
LLMRank~\cite{hou2024large} is the recommendation method that utilizes pre-trained LLMs without additional training or fine-tuning, making it a suitable baseline. It describes three approaches for LLM-based recommendation: Sequential, Recency, and In-context learning (ICL). We compare our method with all three approaches and demonstrate the superiority of PURE when these techniques were applied to our framework, further highlighting its effectiveness. In the following, we describe each approach:

\smallskip\smallskip
  \textbf{1) Sequential.} We provide the LLM with instructions, supplying only the user-item interactions and the candidate list. The LLM is then tasked with ranking the items in the candidate list based on the likelihood of being purchased at time step $t$.\\
  \textbf{2) Recency-Focused.} In the \emph{sequential} prompt above, we add an instruction to emphasize the most recently purchased item, specifically the one bought at time step~($t{-}1$). The additional prompt is:  
  \emph{"Note that my most recently purchased item is \{recent item\}."}\\
  \textbf{3) In-Context Learning.} Unlike the previous \emph{sequential} and \emph{recency-focused} prompts, this approach utilizes user-item interactions only up to time step~($t{-}2$) and the recently purchased item at~($t{-}1$). The additional prompt is: \emph{"I've purchased the following products: \{user-item interactions\}, then you should recommend \{recent item\} to me, and now that I've bought \{recent item\}."}

% \smallskip\smallskip
% \noindent \textbf{2) Recency-Focused.}  
% In the \emph{sequential} prompt above, we add an instruction to emphasize the most recently purchased item, specifically the one bought at time step~($t{-}1$). The additional prompt is:  
% \emph{"Note that my most recently purchased item is \{recent item\}."}

% \smallskip\smallskip
% \noindent \textbf{3) In-Context Learning.}  
% Unlike the previous \emph{sequential} and \emph{recency-focused} prompts, this approach utilizes user-item interactions only up to time step~($t{-}2$) and the recently purchased item at~($t{-}1$). The additional prompt is: \emph{"I've purchased the following products: \{user-item interactions\}, then you should recommend \{recent item\} to me, and now that I've bought \{recent item\}."}

\smallskip\smallskip
\noindent \textbf{Evaluation Setting.} To evaluate the performance of PURE, we adopt a continuous sequential recommendation. In this setup, the LLM is tasked with predicting the item a user is most likely to purchase at time step~$t$. The model receives the user's interaction history up to time step~$(t{-}1)$ in chronological order, along with a candidate set comprising one ground-truth item and 19 randomly sampled non-interacted items. Here, time step~$t$ spans from the user's 4\textsuperscript{th} purchase to their final purchase $k$. To reflect the sequential nature of the task, NDCG scores are first aggregated across multiple recommendation sessions for each user and then averaged across all users.

\smallskip\smallskip
\noindent \textbf{Implementation Details.} To perform this framework, we utilize Llama-3.2-3B-Instruct~\cite{touvron2023llama} as the backbone model for all experiments. Due to the inherent nature of generative language models, it is not always guaranteed that the output will follow the desired format in every response. This issue can be mitigated using structured output formats such as JSON or XML. These formats enable us to enforce consistency and completeness in the model’s output by explicitly defining the expected response structure~\cite{hou2024large, bao2023tallrec}. In our implementation, we prompt the LLM to respond using JSON schemas, which improves reliability during post-processing and facilitates automatic evaluation of model outputs.

\subsection{Experimental Results}
\setlength{\tabcolsep}{8pt}
\begin{table}[h]
  \centering
  \small
  \caption{\textbf{Comparison PURE with Baselines.} We evaluate performance under two data settings: using only item interactions and using item interactions augmented with reviews. $\dagger$ indicates customized baselines where review data is naively incorporated into the original prompt templates designed for item interactions only.}
  \begin{tabular}{@{}clcccccccc@{}}
    \toprule
     &  &  \multicolumn{4}{c}{Games} &  \multicolumn{4}{c}{Movies} \\\cmidrule(lr){3-6} \cmidrule(lr){7-10} 
     Data & Method          &    N@1   &    N@5    &    N@10    &   N@20    &   N@1   &    N@5    &    N@10     &   N@20\\ \cmidrule(lr){1-10}
     
     \multirow{3}{*}{\rotatebox{90}{items}} & Sequential & 10.75 & 18.25 & 23.13 & 28.97 & 9.99  & 15.92 & 20.17 & 26.94 \\
     & Recency    & 15.34 & 24.31 & 28.82 & 34.24 & 12.17 & 17.75 & 22.18 & 28.19 \\
     & ICL        & 14.28 & 26.57 & 30.51 & 35.72 & 12.03 & 19.56 & 23.36 & 29.91 \\ \cmidrule(lr){1-10}

     \multirow{6}{*}{\rotatebox{90}{items + reviews}} 
     & Sequential$^\dagger$ & 11.14 & 19.95 & 24.97 & 32.00 & 8.05 & 13.11 & 17.72 & 25.57\\ 
     & Recency$^\dagger$    & 12.19 & 23.64 & 28.37 & 35.35 & 8.54 & 15.78 & 21.31 & 29.21\\
     & ICL$^\dagger$        & 15.11 & 26.34 & 31.25 & 37.39 & 12.24& 22.10 & 27.31 & 34.52\\ \cmidrule(lr){2-10}
     &\textbf{PURE (Sequential)}& 15.06          & 25.71          & 31.08          &       38.28          & 12.59          & 21.33          & 25.96          &           32.21  \\ 
     &\textbf{PURE (Recency)}   & \textbf{18.18} & 28.90          & 33.91          &     40.69          & 13.85          & 21.99          & 26.53          &               33.37  \\
     &\textbf{PURE (ICL)}       & 16.62          & \textbf{29.81} & \textbf{35.60} & \textbf{42.00} & \textbf{15.80} & \textbf{26.32} & \textbf{32.03} & \textbf{38.93}  \\
    \bottomrule
  \end{tabular}
  \vspace{0.1cm}
  \label{table:main}
  \vspace{-0.2cm}
\end{table}

\setlength{\tabcolsep}{2pt}
\begin{table}[h]
\centering
\caption{\textbf{Component-wise study of PURE.} Each configuration varies which data sources (items, reviews) and which PURE components are used (Rec. = Recommendation, Ext. = Extractor, Upd. = Updater), as indicated by \ding{51}. We report N@k scores ($k\in\{1,5,10,20\}$) and average of input token size (|T|) for Recommender.}
  \small
    \begin{tabular}{@{}cccccccccccccccc@{}}
    \toprule
         & \multicolumn{2}{c}{Data} & \multicolumn{3}{c}{Components} & \multicolumn{5}{c}{Games} &  \multicolumn{5}{c}{Movies} \\\cmidrule(lr){2-3} \cmidrule{4-6} \cmidrule(lr){7-11} \cmidrule(lr){12-16} 
    Method    & items & reviews & Rec. & Ext. & Upd.                                       &    N@1   &    N@5    &    N@10    &   N@20   &   $|T|$   &   N@1      &    N@5    &    N@10     &   N@20    &  $|T|$\\ \cmidrule(lr){1-16}
    \multirow{4}{*}{\rotatebox{90}{Sequential}} & \ding{51}&           & \ding{51} &           &            &   10.75   &   18.25   &   23.13   &   28.97   &   245.52   &   9.99   &   15.92   & 20.17   &   26.94   &   243.89   \\ 
                                                & \ding{51}& \ding{51} & \ding{51} &           &            &   11.14   &   19.95   &   24.97   &   32.00   &   29165.17   &   8.05   &   13.11   & 17.72   &   25.57   &   60429.80   \\
                                                & \ding{51}& \ding{51} & \ding{51} & \ding{51} &            &   16.09   &   26.94   &   32.35   &   40.08   &   486.49   &   13.05  &   21.38   & 26.11   &   32.62   &   459.69   \\
                                                & \ding{51}& \ding{51} & \ding{51} & \ding{51} & \ding{51}  &   15.06   &   25.71   &   31.08   &   38.28   &   415.01   &   12.59  &   21.33   & 25.96   &   32.21   &   384.87   \\ \cmidrule(lr){1-16}
    \multirow{4}{*}{\rotatebox{90}{Recency}}    & \ding{51}&           & \ding{51} &           &            &   15.34   &   24.31   &   28.82   &   34.24   &   253.31   &   12.17  &   17.75   & 22.18   &   28.19   &   249.64   \\
                                                & \ding{51}& \ding{51} & \ding{51} &           &            &   12.19   &   23.64   &   28.37   &   35.35   &   29235.16   &   8.54   &   15.78   & 21.31   &   29.21   &   60509.43    \\
                                                & \ding{51}& \ding{51} & \ding{51} & \ding{51} &            &   20.85   &   31.36   &   36.51   &   43.19   &   602.13   &   16.00  &   24.81   & 29.66   &   36.98   &   565.13  \\
                                                & \ding{51}& \ding{51} & \ding{51} & \ding{51} & \ding{51}  &   18.18   &   28.90   &   33.91   &   40.69   &   485.85   &   13.85  &   21.99  & 26.53   &   33.37   &   458.60   \\ \cmidrule(lr){1-16}
    \multirow{4}{*}{\rotatebox{90}{ICL}}        & \ding{51}&           & \ding{51} &           &            &   14.28   &   26.57   &   30.51   &   35.72   &   268.40   &   12.03  &   19.56  & 23.36   &   29.91   &   261.58   \\
                                                & \ding{51}& \ding{51} & \ding{51} &           &            &   15.11   &   26.34   &   31.25   &   37.39   &   29388.72   &   12.24  &   22.10  & 27.31   &   34.52   &   60800.61   \\
                                                & \ding{51}& \ding{51} & \ding{51} &\ding{51}  &            &   19.60   &   32.96   &   38.21   &   44.97   &   803.60   &   16.05  &   27.25  & 33.11   &   40.15   &   867.36   \\
                                                & \ding{51}& \ding{51} & \ding{51} &\ding{51}  & \ding{51}  &   16.62   &   29.81   &   35.60   &   42.00   &   592.48   &   15.80  &   26.32  & 32.03   &   38.93   &   634.02   \\
    \bottomrule
    \end{tabular}
    % \end{adjustbox}
    \vspace{0.1cm}
  \label{table:ablation}
  \vspace{-2em}
\end{table}

\noindent \textbf{Impact of Review Extractor.} 
\autoref{table:main} compares PURE with (1) three baselines solely based on purchased items; (2) modified baselines, marked with $\dagger$, that additionally utilize users' raw reviews. 
The results reveal that baselines that simply combine item interactions with raw reviews show inconsistent performance improvements.
In contrast, PURE, which leverages the review extractor and profile updater, significantly outperforms all baselines. This demonstrates that {processing reviews at three levels, {i.e.}, like, dislike, and key features, is essential for enhancing performance}.

\smallskip\smallskip
\noindent \textbf{Component-wise Study.}
\autoref{table:ablation} shows the ablation study of PURE, where we analyze the impact of reviews (using or not using) and the effect of components (enabling or disabling the review extractor and profile updater). The use of reviews bring high performance gains only when accompanied by Review Extractor (Ext.). This is due to the sharp increase in input tokens (see the |$T$| column of the 2nd and 3rd rows of each method) as the user continues purchases.

Notably, the best recommendation performance is achieved when Profile Updater (Upd.) is disabled (see the 3rd and 4th rows for each method). This suggests that the well-structured context provided by the Review Extractor alone can lead to strong performance when directly concatenated, even without profile updating. However, it may face a challenge, as the number of purchases grows, leading to significant computational overhead. 
Thus, we use Profile Updater (Upd.) to maintain compact user profiles, reducing input token size by 15--20\% with only a slight 1--3\% performance drop. This trade-off underscores the need for Profile Updater for long-term recommendations.

\begin{figure}[h]
    \centering
    \caption{Trade-off between NDCG and input token size.}
    \includegraphics[width=\linewidth]{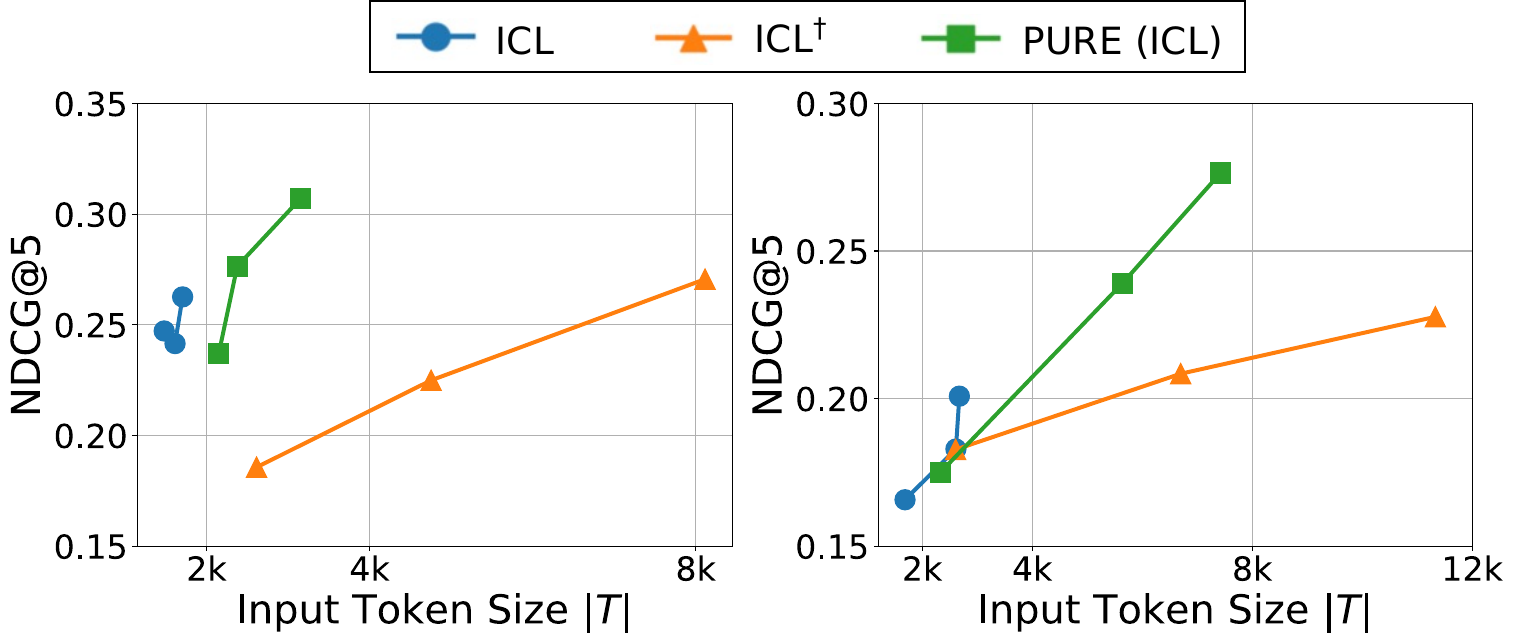}
    \hspace{5em}
    (a) Video Games
    \hspace{15em}
    (b) Movies and TV
    \vspace{-0.5em}
    \label{fig:com}
    \vspace{-0.05cm}
\end{figure}

\smallskip\smallskip
\noindent\textbf{Trade-off Analysis.}
We categorize users into three groups based on the total cumulative review token count per user, as the criterion: 0–500 (short), 500–1000 (middle), and 1000–2000 (long) tokens.
\autoref{fig:com} presents the trade-off between recommendation performance and input token length of the three models including PURE.

PURE achieves the best trade-off, showing the steepest NDCG increase compared to other methods as input token size grows. Therefore, this demonstrates that PURE accurately distills key information from long reviews, while achieving efficiency by minimizing input token growth without information loss, even for long-group users.

\section{Related Works}
\noindent\textbf{Recommendation Setup.} Conventional sequential recommendation methods~\cite{wangsequential, kang2018self, sun2019bert4rec, hidasi2018recurrent, kim2024large} have followed a one-shot prediction setup, in which a user's interaction history is split such that the most recent item is held out as the test set, the second-most recent as the validation set, and the remaining history is used for training. While this setup simplifies the evaluation pipeline, it restricts the model to predicting a single target item, thereby failing to capture the nuanced and evolving nature of user preferences over time. 

\smallskip\smallskip
\noindent\textbf{LLM-based Recommendation.}
A notable example is EXP3RT~\cite{kim2024driven}, which constructs static user profiles by fine-tuning LLMs directly on target recommendation datasets. These fine-tuned models are then used to compute preference scores over candidate items. Tallrec~\cite{bao2023tallrec} proposed the parameter-efficient finetuning (PEFT) method in recommender system, and A-LLMRec~\cite{kim2024large} proposed to finetune the embedding model for LLM to leverage the collaborative knowledge. 
In contrast, train-free models~\cite{wang2023zero} guided users through a conversational process to elicit responses and extract multiple features. 
These features are then used to make personalized recommendations in a conversation-based recommendation framework. 
Also, uncovering ChatGPT's capabilities of recommendation ~\cite{dai2023uncovering} shows ChatGPT is good at reranking the candidates and choosing user preference items while less good at rating. 
InstructRec~\cite{zhang2023recommendation} designed the instruction to recognize the users' intention and preference from context. 

\section{Conclusion}
We propose PURE, a train-free LLM-based recommendation system that operates within a limited input token budget, while maintaining flexibility and eliminating the need for task-specific training. PURE reduces computational costs compared to train-based systems and adapts to various domains. Additionally, we introduce a sequential recommendation task to better model the evolving nature of user preferences over time, moving beyond conventional sequential setups. This work highlights the potential of train-free LLMs in real-world recommendation scenarios.

\noindent \textbf{Limitations.} A notable limitation of our approach is the tendency of the LLM to exhibit hallucination by occasionally recommending items beyond the predefined candidate set, even when explicitly instructed to select from it. This phenomenon underscores the difficulty in imposing strict constraints within LLM-based recommendation models while maintaining flexibility and accuracy. Also, our study was constrained by the inability to utilize datasets containing a larger number of user reviews, which may have provided richer context.

\begin{acknowledgments}
This work was supported by Institute of Information \& communications Technology Planning \& Evaluation (IITP) grant funded by the Korea government (MSIT) (No. RS-2024-00445087). Additionally, this research was supported by the National Research Foundation of Korea (NRF) funded by Ministry of Science and ICT (RS-2022-NR068758).
\end{acknowledgments}

%%
%% Define the bibliography file to be used
\bibliography{sample-ceur}

@article{touvron2023llama,
  title={LLaMA: Open and Efficient Foundation Language Models},
  author={Touvron, Hugo and Lavril, Thibaut and others},
  journal={ArXiv:2302.13971},
  year={2023}
}

@article{dubey2024llama,
  title={The llama 3 herd of models},
  author={Dubey, Abhimanyu and Jauhri, Abhinav and Pandey, Abhinav and Kadian, Abhishek and Al-Dahle, Ahmad and Letman, Aiesha and Mathur, Akhil and Schelten, Alan and Yang, Amy and Fan, Angela and others},
  journal={ArXiv:2407.21783},
  year={2024}
}

@article{achiam2023gpt,
  title= {Gpt-4 technical report},
  author={Achiam, Josh and Adler, Steven and Agarwal, Sandhini and Ahmad, Lama and Akkaya, Ilge and Aleman, Florencia Leoni and Almeida, Diogo and Altenschmidt, Janko and Altman, Sam and Anadkat, Shyamal and others},
  journal={ArXiv:2303.08774},
  year={2023}
}

@article{li2024survey,
  title={A survey on LLM-based multi-agent systems: workflow, infrastructure, and challenges},
  author={Li, Xinyi and Wang, Sai and Zeng, Siqi and Wu, Yu and Yang, Yi},
  journal={Vicinagearth},
  volume={1},
  number={1},
  pages={9},
  year={2024},
  publisher={Springer}
}

@article{ding2024longrope,
  title={Longrope: Extending llm context window beyond 2 million tokens},
  author={Ding, Yiran and Zhang, Li Lyna and Zhang, Chengruidong and Xu, Yuanyuan and Shang, Ning and Xu, Jiahang and Yang, Fan and Yang, Mao},
  journal={ArXiv:2402.13753},
  year={2024}
}

@article{liu2024lost,
  title={Lost in the middle: How language models use long contexts},
  author={Liu, Nelson F and Lin, Kevin and Hewitt, John and Paranjape, Ashwin and Bevilacqua, Michele and Petroni, Fabio and Liang, Percy},
  journal={Transactions of the Association for Computational Linguistics},
  volume={12},
  pages={157--173},
  year={2024},
  publisher={MIT Press One Broadway, 12th Floor, Cambridge, Massachusetts 02142, USA~…}
}

@article{team2024gemma,
  title={Gemma 2: Improving open language models at a practical size},
  author={Team, Gemma and Riviere, Morgane and Pathak, Shreya and Sessa, Pier Giuseppe and Hardin, Cassidy and Bhupatiraju, Surya and Hussenot, L{\'e}onard and Mesnard, Thomas and Shahriari, Bobak and Ram{\'e}, Alexandre and others},
  journal={ArXiv:2408.00118},
  year={2024}
}

@inproceedings{lewis2019bart,
  title={Bart: Denoising sequence-to-sequence pre-training for natural language generation, translation, and comprehension},
  author={Lewis, Mark and Liu, Yinhan and others},
  booktitle={Proceedings of the Annual Meeting of the Association for Computational Linguistics (ACL)},
  year={2020}
}

@inproceedings{karpukhin2020dense,
  title={Dense Passage Retrieval for Open-Domain Question Answering},
  author={Karpukhin, Vladimir and Oguz, Barlas and Min, Sewon and Lewis, Patrick and Wu, Ledell and Edunov, Sergey and Chen, Danqi and Yih, Wen-tau},
  booktitle={Proceedings of Empirical Methods in Natural Language Processing (EMNLP)},
  pages={6769--6781},
  year={2020}
}

@inproceedings{brown2020language,
  title={Language Models are Few-Shot Learners},
  author={Brown, Tom B and Mann, Benjamin and Ryder, Nick and others},
  booktitle={Proceedings of Advances in Neural Information Processing Systems (NeurIPS)},
  volume={33},
  pages={1877--1901},
  year={2020}
}

@inproceedings{lewis2020retrieval,
  title={Retrieval-augmented generation for knowledge-intensive nlp tasks},
  author={Lewis, Patrick and Perez, Ethan and Piktus, Aleksandra and Petroni, Fabio and Karpukhin, Vladimir and Goyal, Naman and K{\"u}ttler, Heinrich and Lewis, Mike and Yih, Wen-tau and Rockt{\"a}schel, Tim and others},
  booktitle={Proceedings of Advances in Neural Information Processing Systems (NeurIPS)},
  volume={33},
  pages={9459--9474},
  year={2020}
}

@inproceedings{hou2024large,
  title={Large language models are zero-shot rankers for recommender systems},
  author={Hou, Yupeng and Zhang, Junjie and Lin, Zihan and Lu, Hongyu and Xie, Ruobing and McAuley, Julian and Zhao, Wayne Xin},
  booktitle={Proceedings of European Conference on Information Retrieval},
  pages={364--381},
  year={2024},
  organization={Springer}
}

@article{wang2023zero,
  title={Zero-shot next-item recommendation using large pretrained language models},
  author={Wang, Lei and Lim, Ee-Peng},
  journal={ArXiv:2304.03153},
  year={2023}
}

@inproceedings{dai2023uncovering,
  title={Uncovering chatgpt’s capabilities in recommender systems},
  author={Dai, Sunhao and Shao, Ninglu and Zhao, Haiyuan and Yu, Weijie and Si, Zihua and Xu, Chen and Sun, Zhongxiang and Zhang, Xiao and Xu, Jun},
  booktitle={Proceedings of the 17th ACM Conference on Recommender Systems},
  pages={1126--1132},
  year={2023}
}

@article{zhang2023recommendation,
  title={Recommendation as instruction following: A large language model empowered recommendation approach},
  author={Zhang, Junjie and Xie, Ruobing and Hou, Yupeng and Zhao, Xin and Lin, Leyu and Wen, Ji-Rong},
  journal={ACM Transactions on Information Systems},
  year={2023},
  publisher={ACM New York, NY}
}

@inproceedings{sun2019bert4rec,
  title={BERT4Rec: Sequential recommendation with bidirectional encoder representations from transformer},
  author={Sun, Fei and Liu, Jun and Wu, Jian and Pei, Changhua and Lin, Xiao and Ou, Wenwu and Jiang, Peng},
  booktitle={Proceedings of the 28th ACM international conference on information and knowledge management},
  pages={1441--1450},
  year={2019}
}

@inproceedings{kang2018self,
  title={Self-attentive sequential recommendation},
  author={Kang, Wang-Cheng and McAuley, Julian},
  booktitle={Proceedings of International Conference on Data Mining (ICDM)},
  pages={197--206},
  year={2018},
  organization={IEEE}
}

@inproceedings{he2017neural,
  title={Neural collaborative filtering},
  author={He, Xiangnan and Liao, Lizi and Zhang, Hanwang and Nie, Liqiang and Hu, Xia and Chua, Tat-Seng},
  booktitle={Proceedings of International Conference on World Wide Web (WWW)},
  pages={173--182},
  year={2017}
}

@inproceedings{he2020lightgcn,
  title={Lightgcn: Simplifying and powering graph convolution network for recommendation},
  author={He, Xiangnan and Deng, Kuan and Wang, Xiang and Li, Yan and Zhang, Yongdong and Wang, Meng},
  booktitle={Proceedings of international ACM SIGIR conference on research and development in Information Retrieval},
  pages={639--648},
  year={2020}
}

@inproceedings{hidasi2018recurrent,
  title={Recurrent neural networks with top-k gains for session-based recommendations},
  author={Hidasi, Bal{\'a}zs and Karatzoglou, Alexandros},
  booktitle={Proceedings of the 27th ACM international conference on information and knowledge management},
  pages={843--852},
  year={2018}
}

@inproceedings{zhong2024memorybank,
  title={Memorybank: Enhancing large language models with long-term memory},
  author={Zhong, Wanjun and Guo, Lianghong and Gao, Qiqi and Ye, He and Wang, Yanlin},
  booktitle={Proceedings of the Association for the Advancement of Artificial Intelligence (AAAI)},
  volume={38},
  number={17},
  pages={19724--19731},
  year={2024}
}

@inproceedings{wangsequential,
  title={Sequential Recommender Systems: Challenges, Progress and Prospects},
  author={Wang, Shoujin and Hu, Liang and Wang, Yan and Cao, Longbing and Sheng, Quan Z and Orgun, Mehmet},
  booktitle={Proceedings of International Joint Conference on Artificial Intelligence Organization (IJCAI)},
  year={2019},
  pages={6332-6338}
}

@inproceedings{ni2019justifying,
  title={Justifying recommendations using distantly-labeled reviews and fine-grained aspects},
  author={Ni, Jianmo and Li, Jiacheng and McAuley, Julian},
  booktitle={Proceedings of the 2019 conference on empirical methods in natural language processing and the 9th international joint conference on natural language processing (EMNLP-IJCNLP)},
  pages={188--197},
  year={2019}
}

@inproceedings{bao2023tallrec,
  title={Tallrec: An effective and efficient tuning framework to align large language model with recommendation},
  author={Bao, Keqin and Zhang, Jizhi and Zhang, Yang and Wang, Wenjie and Feng, Fuli and He, Xiangnan},
  booktitle={Proceedings of the 17th ACM Conference on Recommender Systems},
  pages={1007--1014},
  year={2023}
}

@inproceedings{kim2024large,
  title={Large language models meet collaborative filtering: An efficient all-round llm-based recommender system},
  author={Kim, Sein and Kang, Hongseok and Choi, Seungyoon and Kim, Donghyun and Yang, Minchul and Park, Chanyoung},
  booktitle={Proceedings of the 30th ACM SIGKDD Conference on Knowledge Discovery and Data Mining},
  pages={1395--1406},
  year={2024}
}

@inproceedings{wei2024llmrec,
  title={Llmrec: Large language models with graph augmentation for recommendation},
  author={Wei, Wei and Ren, Xubin and Tang, Jiabin and Wang, Qinyong and Su, Lixin and Cheng, Suqi and Wang, Junfeng and Yin, Dawei and Huang, Chao},
  booktitle={Proceedings of the ACM International Conference on Web Search and Data Mining},
  pages={806--815},
  year={2024}
}

@inproceedings{ren2024representation,
  title={Representation learning with large language models for recommendation},
  author={Ren, Xubin and Wei, Wei and Xia, Lianghao and Su, Lixin and Cheng, Suqi and Wang, Junfeng and Yin, Dawei and Huang, Chao},
  booktitle={Proceedings of the ACM on Web Conference},
  pages={3464--3475},
  year={2024}
}

@inproceedings{zhai2023knowledge,
  title={Knowledge prompt-tuning for sequential recommendation},
  author={Zhai, Jianyang and Zheng, Xiawu and Wang, Chang-Dong and Li, Hui and Tian, Yonghong},
  booktitle={Proceedings of ACM International Conference on Multimedia},
  pages={6451--6461},
  year={2023}
}

@inproceedings{he2023large,
  title={Large language models as zero-shot conversational recommenders},
  author={He, Zhankui and Xie, Zhouhang and Jha, Rahul and Steck, Harald and Liang, Dawen and Feng, Yesu and Majumder, Bodhisattwa Prasad and Kallus, Nathan and McAuley, Julian},
  booktitle={Proceedings of the ACM International Conference on Information and Knowledge Management (CIKM)},
  pages={720--730},
  year={2023}
}

@article{kim2024driven,
  title={driven Personalized Preference Reasoning with Large Language Models for Recommendation},
  author={Kim, Jieyong and Kim, Hyunseo and Cho, Hyunjin and Kang, SeongKu and Chang, Buru and Yeo, Jinyoung and Lee, Dongha},
  journal={arXiv preprint arXiv:2408.06276},
  year={2024}
}


\end{document}